\documentclass[runningheads]{llncs}

\usepackage[mobile]{eccv}

\usepackage{eccvabbrv}

\usepackage{graphicx}
\usepackage{booktabs}
\usepackage{multirow}

\usepackage[accsupp]{axessibility}  

\usepackage[pagebackref,breaklinks,colorlinks,citecolor=eccvblue]{hyperref}

\usepackage{orcidlink}

\usepackage{xcolor, colortbl}
\usepackage{bbm}
\usepackage{pifont}
\usepackage{nicefrac, xfrac}

\definecolor{lblue}{rgb}{0.1, 0.2, 0.7}
\definecolor{orange}{rgb}{1.0, 0.5, 0.0}
\definecolor{purple}{rgb}{0.5, 0.2, 0.9}
\definecolor{Gray}{gray}{0.9}
\definecolor{plus}{HTML}{0071bc}
\definecolor{minus}{RGB}{153,10,10}
\definecolor{darkgreen}{RGB}{56,87,35}
\definecolor{darkred}{RGB}{192,0,0}
\newcommand{\up}{\bf \fontsize{8}{42} \color{plus}{$\uparrow$}}
\newcommand{\down}{\bf \fontsize{8}{42} \color{minus}{$\downarrow$}}
\newcommand{\cmark}{\ding{51}}
\newcommand{\xmark}{\ding{55}}
\newcommand{\expnum}[2]{{#1}\mathrm{e}{-#2}}

\newcommand{\netnameshort}{Flaming-Net}
\definecolor{minus}{RGB}{153,10,10}
\definecolor{darkgreen}{RGB}{56,87,35}
\definecolor{darkred}{RGB}{192,0,0}

\newlength\secmargin
\newlength\subsecmargin
\newlength\subsubsecmargin
\newlength\paramargin
\newlength\abovetabcapmargin
\newlength\belowtabcapmargin
\newlength\abovefigcapmargin
\newlength\belowfigcapmargin
\begin{document}

\title{Flow-Assisted Motion Learning Network for Weakly-Supervised Group Activity Recognition} 

\titlerunning{Flow-Assisted Motion Learning Network for WSGAR}

\author{Muhammad Adi Nugroho\orcidlink{0000-0002-9360-5441} \and
Sangmin Woo\orcidlink{0000-0003-4451-9675} \and
Sumin Lee \and
Jinyoung Park \and
Yooseung Wang \and
Donguk Kim \and
Changick Kim\orcidlink{0000-0001-9323-8488}}

\authorrunning{M.A.~Nugroho et al.}

\institute{\vspace{-1mm} Korea Advanced Institute of Science and Technology (KAIST) \vspace{1mm}\\
\fontsize{7.6}{9}\selectfont{\email{\{madin,~smwoo95,~suminlee94,~jinyoungpark,~yswang,~kdu3613,~changick\}@kaist.ac.kr}}}

\maketitle

\vspace{-5mm}
\begin{abstract}
  Weakly-Supervised Group Activity Recognition (WSGAR) aims to understand the activity performed together by a group of individuals with the video-level label and without actor-level labels. We propose \textbf{Fl}ow-\textbf{A}ssisted \textbf{M}otion Learn\textbf{ing} Network (\textbf{\netnameshort}) for WSGAR, which consists of the motion-aware actor encoder to extract actor features and the two-pathways relation module to infer the interaction among actors and their activity. \netnameshort~leverages an additional optical flow modality in the training stage to enhance its motion awareness when finding locally active actors. The first pathway of the relation module, the actor-centric path, initially captures the temporal dynamics of individual actors and then constructs inter-actor relationships. In parallel, the group-centric path starts by building spatial connections between actors within the same timeframe and then captures simultaneous spatio-temporal dynamics among them. We demonstrate that \netnameshort~ achieves new state-of-the-art WSGAR results on two benchmarks, including a 2.8\%p higher MPCA score on the NBA dataset. Importantly, we use the optical flow modality only for training and not for inference.
\end{abstract}

\section{Introduction}

\label{sec:intro}
The Group Activity Recognition (GAR) task involves classifying the collective activity performed by a group of individuals within a given video. This area has garnered growing interest owing to its diverse applications, spanning sports video analysis, video surveillance, and comprehension of social scenes. GAR introduces new challenges in human action understanding due to its distinction from the primarily studied individual or paired action recognition~\cite{Lin2019-tsm,Wang-TSN,Liu2022-videoswin}. It involves complex spatio-temporal relationships among multiple actors in a long temporal sequence. GAR is studied initially in a fully supervised setting, where individual bounding boxes and actor actions are available. However, these annotations are costly to acquire, both for the dataset creation and for the inference stage. This challenge motivates the research area of Weakly-Supervised Group Activity Recognition (WSGAR), where only the group activity label is available.

\begin{figure*}[t]
\begin{center}
\includegraphics[width=1.0\linewidth]{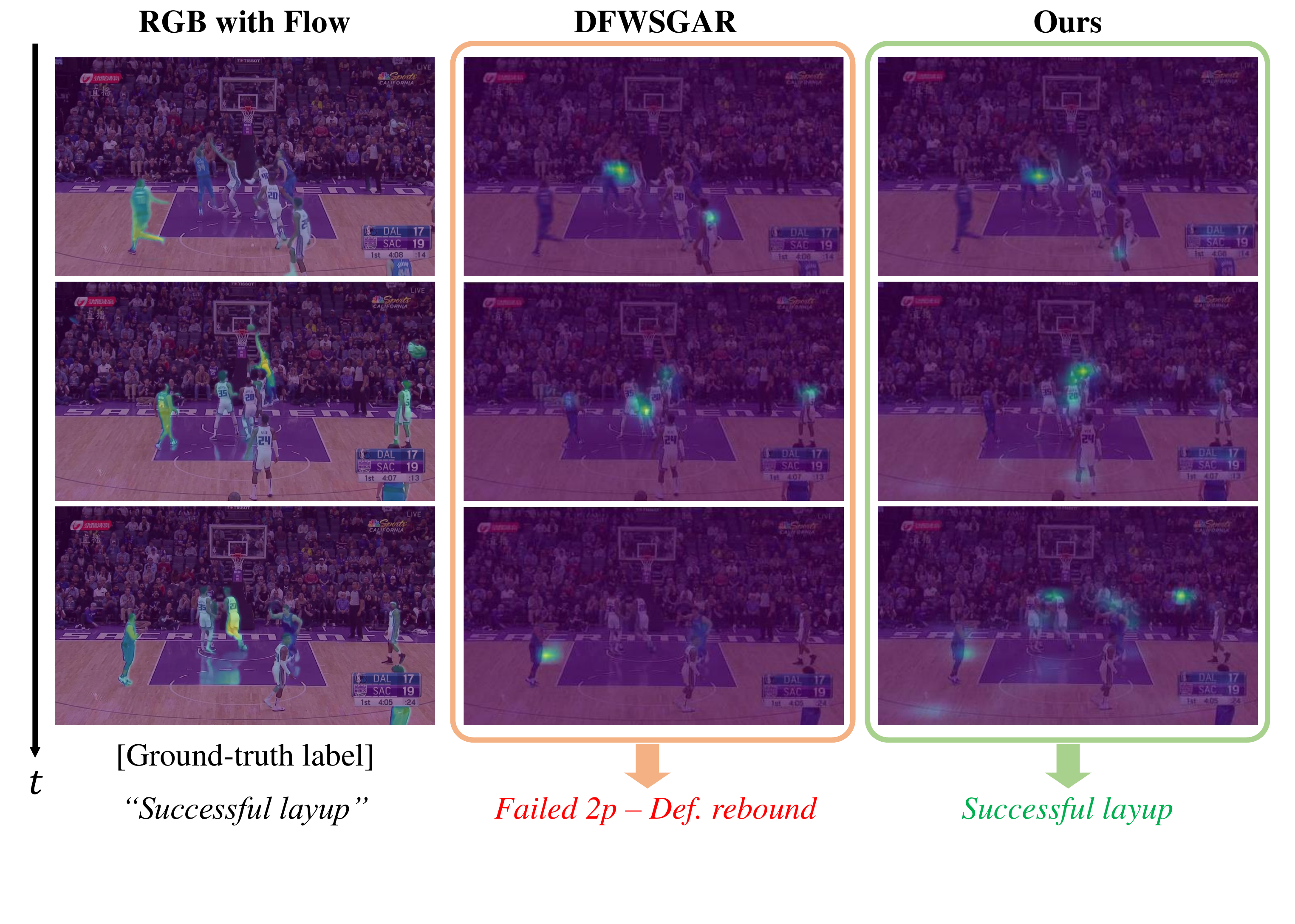}
\end{center}
\vspace{-15mm}
\caption{ 
\textbf{Importance of motion-awareness.} In WSGAR, it is crucial to capture the movements of key actors. In the second frame, DFWSGAR~\cite{Kim2022-dfwsgar} fails to recognize the layup action, as the player doing the layup is not strongly highlighted. In contrast, \netnameshort~strongly highlights the player performing the layup, thanks to its motion awareness. \netnameshort~leverages optical flow as a learning guidance motivated by how an area with intense optical flow indicates the existence of key actors, as seen in the second frame. Lastly, in the third frame, our method successfully identifies multiple key actors, such as the celebrating scoring team, the defending team looking down, and the referee, as strong cues to correctly predict a successful layup attempt.
}
\label{fig:fig1}
\vspace{-8mm}
\end{figure*}

WSGAR poses a significant challenge in acquiring the actor features, requiring the model to learn distinctive cues that help determine the location or feature that highly correlates with key actors. Flow information is one of the useful clues that highlight the key actors. In \cref{fig:fig1}, we show a clipped sequence of a successful layup activity in a basketball game. The leftmost column demonstrates how adding flow information strongly accentuates the key actors, especially the player performing the layup and defending players. Unfortunately, multiple non-important players are also highlighted. DFWSGAR~\cite{Kim2022-dfwsgar}, a state-of-the-art WSGAR model, more selectively highlights the important player compared to flow information, reducing needless noises. However, the attention to players in the under-ring area is also reduced, causing a loss of detailed information and resulting in an incorrect prediction. We propose the \textbf{Fl}ow-\textbf{A}ssisted \textbf{M}otion Learn\textbf{ing} Network (\textbf{\netnameshort}) for WSGAR, in which one of the key work principles is to selectively highlight key actors while also paying high attention to the crucial, highly active areas. The attention maps of our model, shown in the third column, show how it focuses on the under-ring area in the second frame, especially on actively moving players. Additionally, the third frame also shows how the model highlights supporting key actors, resulting in correct prediction.

\begin{figure*}[!t]
\begin{center}
\includegraphics[width=1.0\linewidth]{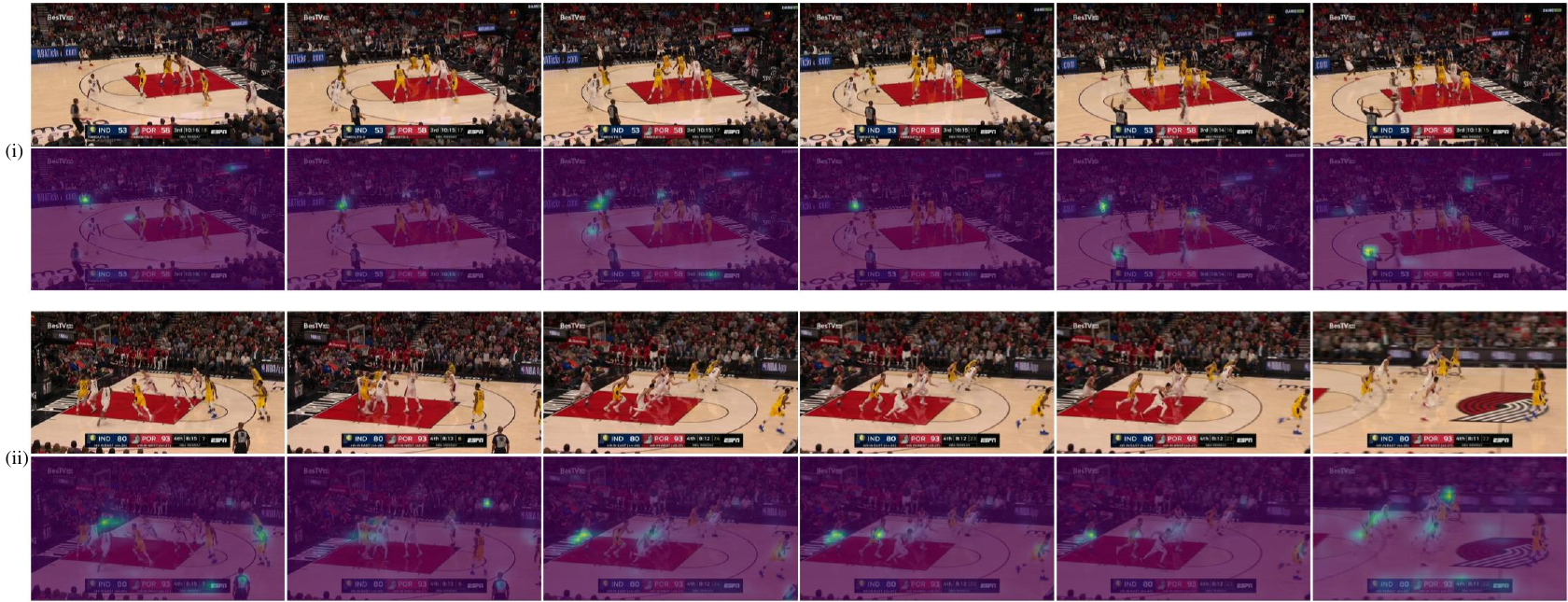}
\end{center}
\vspace{-6mm}
\caption{\textbf{Visualization of the motion-aware actor encoder attention maps} on the activity of (i) \textit{3p-success} and (ii) \textit{2p-layup-fail.-def}. \netnameshort~highlights the key actors, \textit{e.g.} 3-point shooter and referee giving hand sign in the sequence (i), and offensive player doing layup and counter-attacking defensive players in (ii).}
\label{fig:fig5}
\vspace{-5mm}
\end{figure*}

The second key principle of our model is combining powerful cues from a long temporal dynamic of individual actors with multi-actor interactions in a short time window when inferring video-level inter-actor relationships. Figure~\ref{fig:fig5} shows a longer clip of (i) \textit{3p-success} and (ii) \textit{2p-failed-def. rebound} activity sequences. In Fig.~\ref{fig:fig5}-(i), the long-range individual temporal dynamic of the 3p-shooter is crucial to recognize the activity. In the last two frames, the referee and the ring are also highlighted as the points of focus, due to their importance in determining whether the 3-point attempt is successful or not. In Fig.~\ref{fig:fig5}-(ii), it is more crucial to determine the interaction among actors in a short time window, particularly in frames 1 and 2 where the players are hustling to catch the ball. The progressing sequence also shows how multiple actors interact together, e.g. players on the same team running to the opposition side. The spatio-temporal relation module combines the local dynamics of an actor-centric approach with the comprehensive global motion-oriented approach to establish spatio-temporal relationships among actors with parallel aggregation paths.

We evaluate the proposed framework on two datasets, Volleyball~\cite{Ibrahim2016-volleyball} and NBA~\cite{Yan2020-SAM}. Our framework achieves competitive performance with state-of-the-art methods on these two benchmarks in the weakly supervised learning setting compared with both fully supervised and weakly supervised models. The contribution of this paper is threefold:
\begin{itemize}
\item[$\bullet$] We propose a novel learning method guided by optical flow frames and an auxiliary temporal contrastive loss to build motion-awareness when extracting key actor features without bounding box annotations. Note that optical flow is not required in the inference stage.
\item[$\bullet$] We build a spatio-temporal actor relationship module to infer collective activity containing two parallel paths: (i) actor motion path that models activity as an interaction between sequence long individual long-range temporal dynamic, and (ii) group motion path that models activity as progressing spatio-temporal connections between one or multiple groups of actors.
\item[$\bullet$] Our method achieves state-of-the-art performance on two benchmark datasets while maintaining the computational costs and not requiring expensive ground truth bounding boxes or individual action annotations.
\end{itemize}

\section{Related Work}
\subsection{Group Activity Recognition}\vspace{\subsecmargin}
Group activity recognition requires modeling the relationships between multiple actors. Typically, individual actor features are extracted from the image feature map using methods such as RoI-Align~\cite{He2017-maskrcnn} or RoI-Pooling~\cite{Ren2015-RoiPool}, with the help of bounding boxes. Then, the actor features are formed into a group-level representation by constructing spatio-temporal relationships among actor features. Earlier methods utilize hand-crafted features with probabilistic graphical models~\cite{Amer2014-HiRF,Choi2012-AUnified} or AND-OR graphs~\cite{Amer2012-CostSensitive,Shu2015-JointInference} to model the connections between individual actions and group activity. Some other works employ multi-stage RNNs to encode the hierarchical temporal structure of connections inside the group~\cite{Bagautdinov2017-social,Ibrahim2016-volleyball,Deng2016-Structure,Shu2017-CERN}. Alternatively, the RNN constructs a graphical structure with actor features as its element~\cite{Deng2016-Structure,Qi2020-StagNET}. RNNs such as LSTM capture individual action dynamics and aggregate individual information to infer group activity. 

Recent approaches show trend towards adopting relational modeling~\cite{Azar2019-ConvGAR} to model relationships between actors, including the widely used graph-based approaches~\cite{Ehsanpour2020-JointLearning, Hu2020-Progressive,Wu2019-ARG,Yuan2021-DIN,Yan2020-higcin}, where actor features are assigned as nodes and their relationships are represented as edges. Graph-based methods subsequently employ the way to evolve relation graphs, such as Graph Convolutional Network (GCN)~\cite{Wu2019-ARG} or Graph Attention Network (GAT)~\cite{Ehsanpour2020-JointLearning}. Some approaches build hierarchical relationships from body-level representation to group-level representation by constructing a cross-inference module to embed spatio-temporal features~\cite{Yan2020-higcin} or by utilizing dynamic relation and dynamic walk offsets to build person-specific interaction graph~\cite{Yuan2021-DIN}. On the other hand, Azar et al.~\cite{Azar2019-ConvGAR} introduce the notion of activity map to encode spatial relations between individuals.

Transformer-based methods~\cite{Gavrilyuk2020-ATGAR, Li2021-Groupformer, Chappa2023-SPARTAN,Tamura2022-Hunting,Wu2023-active} have emerged as effective approaches for modeling the relationships between actors. These methods typically employ a transformer module on top of the actor features, such as SACRF~\cite{Pramono2020-SACRF} which embed spatio-temporal relational contexts with conditional random fields (CRF) or joint spatio-temporal contexts regarding intra- and inter-group relations\cite{Li2021-Groupformer}. Yuan~\etal\cite{Yuan2021-LearningVisualGAR} encode person-specific scene contexts per individual feature. Tamura~\etal\cite{Tamura2022-Hunting} combine social group detection and group activity classification with a region-aware transformer to identify and aggregate features relevant to social group activities. Dual-AI~\cite{Han2022-dualAI} construct two complementary spatio-temporal views to infer relations between actors. Wu~\etal\cite{Wu2023-active} use the transformer as a hierarchical relation inference network to process multi-scale feature maps based on active spatial positions.

Our work differs from the aforementioned approaches in that they are primarily designed for the fully supervised setting. In contrast, we focus on the challenging weakly supervised setting, where obtaining the features of key actors is more difficult due to the lack of any actor-level annotations. Despite this difference, we draw inspiration from common practices found in state-of-the-art fully supervised GAR methods, particularly in using an inter-actor relation module to build video-level representation and two parallel aggregation paths.

\subsection{Weakly-Supervised Group Activity Recognition}\vspace{\subsecmargin} 

The goal of WSGAR is to reduce the labeling cost associated with detailed actor-level annotations, such as actor bounding boxes and action classes. Yan~\etal\cite{Yan2020-SAM} introduce the NBA dataset for WSGAR task, along with a specialized WSGAR model called SAM. SAM utilizes an off-the-shelf object detector to provide bounding box annotations, followed by a relation graph to prune noisy outputs from the detector. SAM has a drawback of reliance on an object detector not directly optimized for the GAR problem. To overcome this, Kim~\etal\cite{Kim2022-dfwsgar} propose a detector-free network based on a transformer decoder. Similar to DETR~\cite{Carion2020-DETR}, they use the transformer to find relevant actors using learnable queries and a cross-attention decoder. The identified actors are then transformed into tokens and processed with a temporal convolution encoder and actor relation self-attention module. Du~\etal\cite{Du2023-LRMMGCM} address the information loss in the encoding process by using a decoupled 3D convolutional network intended to better separate camera motion and actor motions. Several methods employ a pretraining self-supervision strategy, for example, Chappa~\etal\cite{Chappa2023-SPARTAN} adopt the video transformer for activity recognition with a self-supervised pretraining strategy. Con-RPM~\cite{Zhou2023-contextualized} uses the transformer architecture for self-supervised GAR by modeling high-level semantic patterns and learning how to predict future group relations.

Compared to all these works, our method combines the strengths of actor-focused methods and group motion learning approaches to build relationships within groups.
We leverage strong actor features for robust group relationship modeling and incorporate group motion learning, where motions are built from spatio-temporal local connections. We further enhance our model by adding an extrinsic objective of local motion learning, guided by an additional optical flow modality. Groupformer~\cite{Li2021-Groupformer} also uses flow modality by concatenating the flow-extracted feature to the actor feature to enhance the actor relation network. SBGAR~\cite{li2017-sbgar} combines RGB and flow features for semantic-based action and activity recognition. However, unlike these methods, we do not use flow in the inference stage. Instead, we utilize it as guidance in the learning stage, in a straightforward manner, without the need for an additional extraction network.
\begin{figure*}[!t]
\begin{center}
\includegraphics[width=1.0\linewidth]{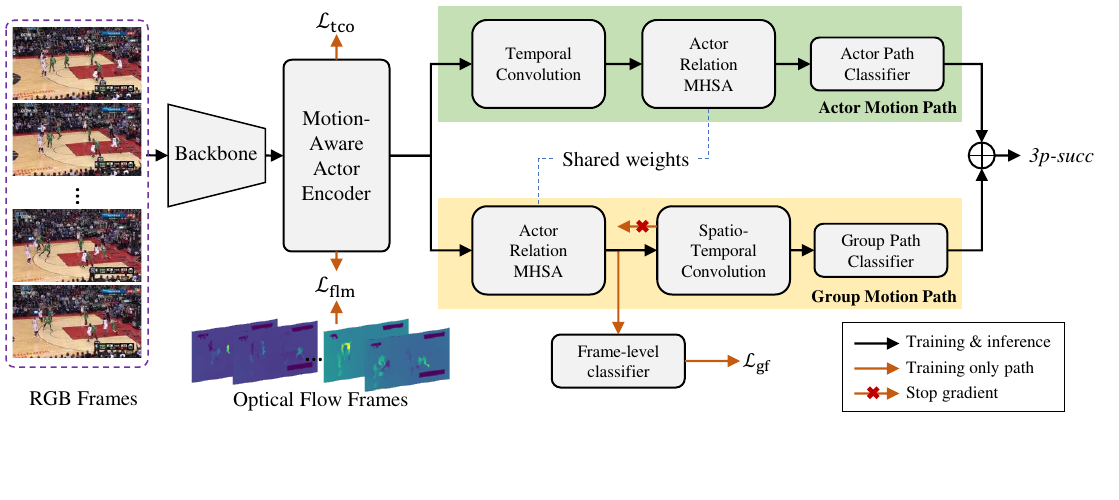}
\end{center}
\vspace{-16mm}
\caption{
 \textbf{Overall architecture of Flaming-Net}. For each frame, a 2D CNN backbone generates a feature map and the motion-aware actor encoder extracts actor features that represent important actors or entities. The sequence of actor features is then forwarded to two different paths: actor motion path and group motion path. To improve the learning process, we add local motion learning with a flow-map-based label to help the model learn local motion. Additionally, we include a temporal consistency loss and a frame-level classifier. Note that optical flow is only used in the training stage.
}
\label{fig:fig2}
\vspace{-6mm}
\end{figure*}

\vspace{-6mm}
\section{Flow-Assisted Motion Learning Network}
\vspace{\secmargin}
The major challenge of WSGAR is finding the key actors in the video. Interestingly, they tend to be more active than their surroundings in an activity sequence, so we can use visual cues associated with motion to find them. \netnameshort~is built to exploit this knowledge of individual actor motion and interactive multi-actor motion to infer group activity. The overview of the network is presented in \cref{fig:fig2}. The backbone network firstly extracts RGB frames into feature maps, $\textbf{F}_0\in\mathbb{R}^{T\times H\times W \times C_0}$. Then, they are processed frame-per-frame through a motion-aware actor encoder to extract a set of actor features. We incorporate the knowledge from the optical flow map into the encoder using a contrastive loss, along with a temporal consistency objective to enhance the encoder motion awareness.  Following this, the actor features are processed through a relation module consisting of the actor motion path and the group motion path.

\subsection{Motion-Aware Actor Encoder}\vspace{\subsecmargin}

\begin{figure}[t]
\begin{center}
\includegraphics[width=1.0\columnwidth]{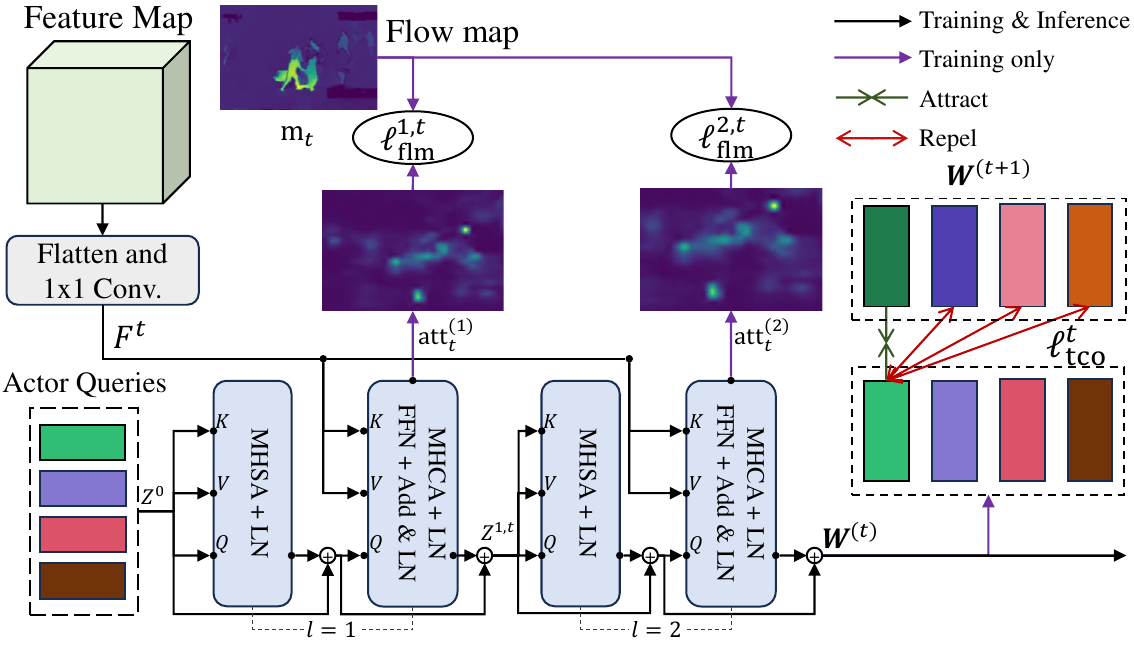}
\end{center}
\vspace{-10mm}
\caption{
\textbf{Motion-aware actor encoder} transforms actor queries and the feature map using a series of multi-head attention modules to generate actor features. The flow learning loss guides the encoders using the optical flow map to highlight active key actors. The temporal consistency loss $\ell_{\text{tco}}$ encourages each token to represent the same actor across frames. The loss \textcolor{darkgreen!70!green}{\textbf{attracts}} temporally adjacent actor features belonging to the same index, while \textcolor{darkred}{\textbf{repels}} those with different indices. Here, we assume $L=2$.
}
\label{fig:fig3}
\vspace{-6mm}
\end{figure}

Motion-aware actor encoder transforms a set of learnable actor queries and the extracted feature map into actor features frame-per-frame. The encoder consists of $L$ layer stacks, each containing a multi-head self-attention (MHSA) and a multi-head cross-attention (MHCA) as visualized in \cref{fig:fig3}. In the layer $l$, the MHCA \textit{Key} and \textit{Value} are the flattened and compressed feature map at frame $t$, $F^t\in\mathbb{R}^{HW \times C}$, while the \textit{Query} is the sum of two elements: (1) the processed actor queries from previous layer $Z^{l-1}\in\mathbb{R}^{K\times C}$, and (2) the output of the MHSA applied to $Z^{l-1}$. The encoder attention mechanism localizes and encodes actors from the given feature maps into tokens. For $T$ frames, the encoder output $\textbf{W}^{(t)} = \{\textbf{w}_i^{(t)}\}_{i=1}^{K}$ is stacked to form $\textbf{W} = [\textbf{W}^{(1)}, \textbf{W}^{(2)}, \ldots, \textbf{W}^{(T)}] \in \mathbb{R}^{T \times K \times C}$, the sequence of action features in form of a bag of $T \times K$ token embeddings of dimension $C$. Note that $\textbf{w}_i^{(t)}$ denotes the $i$-th token embedding for the frame $t$.

\vspace{\subsubsecmargin}\subsubsection{Flow-guided learning.} The actor encoder ideally attends more to the areas with relevant actors. To achieve this, we introduce an auxiliary learning objective using extra guidance of optical flow as it can capture strong cues of actor motion in a frame-per-frame manner. We align the magnitude of optical flow with the attention map of the encoder block. In an action encoder block, the attention map value is the average of the softmax result from the query and key multiplication. Every block generates $K$ attention maps as there are $K$ actor queries. Then, these attention maps are averaged to representative attention map value to be aligned, $\textrm{att}_t(l)$, for encoder block $l$ on $t$ time frame. However, not all relevant key actors contain strong flow magnitude, so we only averaged across the first $K_{\text{flm}}$ number of attention maps and let other attention maps focus on important yet low flow value areas.

We formulate the flow learning loss ($\mathcal{L}_{\text{flm}}$) to align the attention and the optical flow maps, \textbf{m}. Firstly, we reshape the attention and the flow maps into $\mathbb{R}^{NT\times HW}$ vectors. Then, we pair the attention and optical flow map by assigning maps with the same time frame as the positive pairs and those with different time frames as the negative pairs. Then, for each MHCA block, we calculate the contrastive loss and accumulate them over $L$ stack as expressed in \cref{eq:loss_flow},
\vspace{-1mm}
\begin{equation}
    \label{eq:loss_alignment_p1}
    \begin{aligned}
    \begin{split}
 &h(u,v)=\exp((\text{sim}(u,v)/\tau), \: \;
 &\text{sim}(u,v)=u^\intercal v/(\|u\|\|v\|), &
\end{split}
    \end{aligned}
\end{equation}
\vspace{-5mm}
\begin{equation}
    \label{eq:loss_alignment}
    \begin{aligned}
    \ell_{\mbox{\scriptsize flm}}(\textbf{att}^{(l)},\textbf{m}) = -\frac{1}{NT}\sum\limits_{i=1}^{N T}\mathrm{log}\frac{h(\text{att}_i^{(l)},\text{m}_i)}{\sum\limits_{j=1}^{NT}\mathbbm{1}_{i\neq j}h(\text{att}^{(l)}_i,\text{m}_j)},
    \end{aligned}
\end{equation}
\vspace{-5mm}
\begin{equation}
    \label{eq:loss_flow}
    \begin{aligned}
    \mathcal{L}_{\text{flm}}=\frac{1}{L} \sum\limits_{l=1}^{L}\ell_{\mbox{\scriptsize flm}}(\textbf{att}^{(l)},\textbf{m}),
    \end{aligned}
\vspace{-2mm}
\end{equation}
where $l$ is the index of motion-aware encoder block, $\tau$ denotes a temperature parameter, $N$ denotes the number of data in one batch, and $\mathbbm{1}_{j\neq i}\in\{0,1\}$ is an indicator function evaluating to 1 if $j \neq i$. To generate the flow map, we use \textit{FlowNet2}~\cite{Ilg2017-flownet2}  in $H_0\times W_0$ resolution, then resize the result to $H\times W$.

\vspace{\subsubsecmargin}\subsubsection{Temporal consistency.} We add auxiliary loss $\mathcal{L}_{\text{tco}}$ to encourage features with the same index across time to represent the same actors or have similar feature values. We achieve this by using bidirectional contrastive loss between temporally adjacent feature tokens. Tokens with the same index constitute the positive pairs, and tokens with different indices constitute the negative pairs.
\vspace{-1mm}
\begin{equation}
    \label{eq:loss_alignment_temp}
    \resizebox{\linewidth}{!}{
        $\ell_{\mbox{\scriptsize tco}}(\textbf{W}^{(t)}) = -\frac{1}{NK}\sum\limits_{i=1}^{NK} \left(\mathrm{log}\frac{h(\textbf{w}_i^{(t)},\textbf{w}_i^{(t+1)})}{\sum\limits_{j=1}^{NK}\mathbbm{1}_{j\neq i}h(\textbf{w}^{(t)}_i,\textbf{w}^{(t+1)}_j)} + \\    \mathrm{log}\frac{h(\textbf{w}_i^{(t+1)},\textbf{w}_i^{(t)})}{\sum\limits_{j=1}^{NK}\mathbbm{1}_{j\neq i}h(\textbf{w}^{(t+1)}_i,\textbf{w}^{(t)}_j)}\right),$
    }
\end{equation}

\vspace{-2.5mm}
\begin{equation}
    \label{eq:loss_alignment_2}
    \begin{aligned}
    \mathcal{L}_{\text{tco}}=\frac{1}{T-1} \sum\limits_{t=1}^{T-1}\ell_{\mbox{\scriptsize tco}}(\textbf{W}^{(t)}).
    \end{aligned}
\vspace{-7mm}
\end{equation}

\begin{figure}[t]
\begin{center}
\includegraphics[width=1.0\columnwidth]{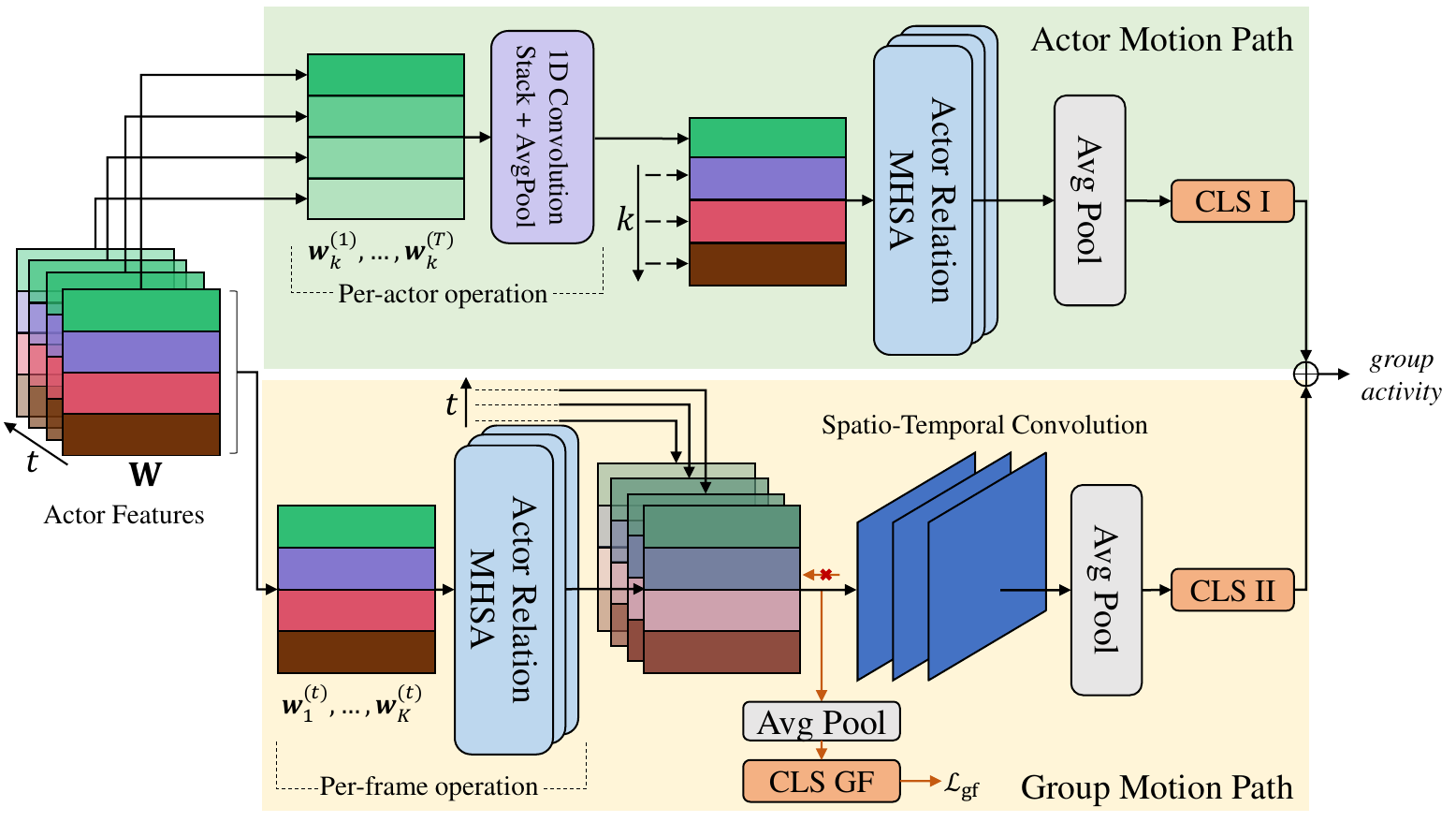}
\end{center}
\vspace{-7mm}
\caption{
\textbf{Actor relation module.} The actor motion path initially generates temporally evolved representations of each actor, then builds inter-actor relations using MHSA. In contrast, the group motion path initially builds inter-actor relationships using MHSA per frame. Then, these features are stacked along the temporal dimension and processed spatio-temporally using a 2D convolution stack. We use separate MLP layer classifiers on the paths and average the outputs of these two classifiers for final prediction.}
\label{fig:fig4}
\vspace{-3mm}
\end{figure}

\subsection{Actor-centric Motion Path}\vspace{\subsecmargin}
Within this path, the model first infers dynamic motion for each actor individually, then builds a group representation by combining them spatially, as shown visually in the upper part of \cref{fig:fig4}. In the first step, the action features $\textbf{W}_{i} = \{\textbf{w}_i^{(t)}\}_{t=1}^{T} \in \mathbb{R}^{T \times C}$ from the $i$-th actor at different frames are fused to form an $i$-th aggregated actor feature $\tilde{\textbf{w}_i} \in \mathbb{R}^{D}$ for all $i$. We use a stack of 1D convolution layers that operate on the temporal dimension to aggregate them. Specifically, we reshape the output token embedding \textbf{W} to $K \times T \times C$ tensor, then feed it to the convolution stack with each layer followed by ReLU\cite{Nair2010-relu} operation. The tokens are then aggregated in the temporal dimension with $AvgPool$ operation, generating a set of $K$ vectors with dimension $C$  for each video. Then, we construct an inter-actor relationship using a single-layer multi-head self-attention. Finally, the actor-path group representation $\textbf{f}_A$ is obtained by applying $AvgPool$ operation over $K$ dimension.

\subsection{Group-centric Motion Path}\vspace{\subsecmargin}
The group-centric motion path facilitates interactions of actor features from different frames, as these represent cues such as multiple actors forming small groups moving together. The lower part of \cref{fig:fig4} illustrates the process, which begins by constructing spatial actor relationships per frame. Specifically, the actor relation MHSA, processes action features on time $t$, $\textbf{W}^{t}$, for each frame. We then reshape the output of this MHSA across frames into $C\times K \times T$ tensor. After that, we process them with a 2D convolution stack, where each convolution layer is activated by ReLU\cite{Nair2010-relu}, to build spatio-temporal relationships across these features. Lastly, we perform $AvgPool$ across the spatio-temporal dimension to obtain the vectorized representation. To improve performance, we add per-frame classification loss $\mathcal{L}_{\text{gf}}$ by aggregating the token features per frame and putting an additional classifier after constructing the spatial relationship with MHSA. We stop the gradient before the 2D convolution stack to improve learning stability.

\subsection{Learning Objective and Inference}\vspace{\subsecmargin}
We use separate MLP classifiers on the actor-centric and group-centric motion paths. The prediction score is obtained by averaging the outputs of these two classifiers. The network is trained end-to-end by summing all the losses, where $\mathcal{L}_{\text{CE}}$ is the cross-entropy loss of the activity prediction, and $\rho$ is the maximum softmax prediction score for each video to adjust the influence of the $\mathcal{L}_{\text{flm}}$. 
\begin{equation}
\mathcal{L}_{\text{sum}}=\mathcal{L}_{\text{CE}}+(1-\rho)\mathcal{L}_{\text{flm}}+\mathcal{L}_{\text{tco}}+\mathcal{L}_{\text{gf}}.
    \label{eq:training_loss}
\end{equation}

\section{Experiments}

\subsection{Datasets}\vspace{\subsecmargin}

\subsubsection{Volleyball dataset.}
This dataset~\cite{Ibrahim2016-volleyball} consists of 4830 clips taken from 55 videos. 3494 clips among them are for training and 1337 clips are for testing.  The center frame of each clip is labeled with the activity label and action label along with the bounding box per player. Bounding box tracklets of players along the 10 frames before and after the center frame are provided by Bagautdinov~\etal~\cite{Bagautdinov2017-social} and serve as the ground truth. In the WSGAR setting, only the group activity labels are available for training. We adopted Multi-class Classification Accuracy (MCA) and Merged MCA for our experiment evaluation.  For computing Merged MCA, we merge the classes \textit{right set} and \textit{right pass} into \textit{right pass-set}, and \textit{left set} and \textit{left pass} into \textit{left pass-set} as in SAM\cite{Yan2020-SAM}.

\vspace{\subsubsecmargin}\subsubsection{NBA dataset.}
This dataset~\cite{Yan2020-SAM} consists of 7624 clips for training and 1548 clips for testing. Only 1 out of 9 group activity labels is provided for each clip, without any individual-level annotation. Since each video clip is 6 seconds long and usually exhibits a nontrivial temporal structure, the dataset requires a model that captures long-term temporal dynamics compared with other GAR benchmarks. Also, it is a challenging benchmark due to fast movement, camera view change, and a varying number of people in each frame. For evaluation, we adopt Multi-class Classification Accuracy (MCA) and Mean Per Class Accuracy (MPCA) metrics; MPCA is adopted due to the class imbalance issue of the dataset.

\subsection{Implementation Details}\vspace{4mm}

\vspace{\subsubsecmargin}\subsubsection{Hyperparameters.} We use an ImageNet pretrained ResNet-18~\cite{he2016-resnet} or Inception-v3~\cite{szegedy2015-inception} with motion augmentation as used in~\cite{Kim2022-dfwsgar} for the backbone. We set $L=6$ with four number of attention heads and $C=128$. The temporal convolution in the actor motion path consists of three 1D convolutional layers with a kernel size of 5 without padding for the NBA dataset and two 1D convolutional layers with a kernel size of 3 with zero-padding are used for Volleyball.  The spatio-temporal convolution consists of two 2D convolution layers with a kernel size of $5\times3$ for NBA and $3\times3$ for Volleyball. We use segment-based sampling~\cite{Wang-TSN} to select $T$ frames from each video in the NBA dataset, and random sampling in the Volleyball dataset. For the NBA dataset, we set $T=18$ and for the Volleyball dataset $T=5$, and we resize each frame to $W_0=1280$ and $H_0=720$.

\vspace{\subsubsecmargin}\subsubsection{Training.} Our model is optimized by ADAM\cite{Kingma2015-Adam} with $\beta_{1}=0.9$, $\beta_{2}=0.999$, and $\epsilon=\expnum{1}{8}$ for 30 epochs. Weight decay is set to $\expnum{1}{4}$ for the NBA dataset. Learning rate is initially set to $\expnum{1}{6}$ with linear warmup to $\expnum{1}{4}$ for 5 epochs, and linearly decayed after the $6^{\textrm{th}}$ epoch. We use a mini-batch of size 4.

\begin{table}[t]
\begin{center}
\caption{\textbf{Comparison with the state-of-the-art GAR models and video backbones on the NBA\cite{Yan2020-SAM} dataset}. Numbers in \textbf{bold} indicate the best performance and \underline{underlined} ones are the second best. '\dag' indicates that the results are from Kim~\etal\cite{Kim2022-dfwsgar} reproduction, and others are as reported from the original paper. ${\ast}$ indicates the method requires additional pre-training before a supervised fine-tuning.
}
\label{table:SOTA_NBA}
\vspace{-3mm}
\resizebox{0.95\columnwidth}{!}{
\setlength{\tabcolsep}{1.5mm}
\begin{tabular}{l c c c c c c}
\hline
Method & Published on & Backbone & \#Params{\down} & FLOPs{\down} & MCA{\up} & MPCA{\up}      \\
\hline
\addlinespace[0.5ex]
\multicolumn{3}{l}{\textbf{Video backbone}} \\
\addlinespace[0.5ex]
\dag~TSM\cite{Lin2019-tsm} & ICCV'19 & Resnet-18 & 11.2M &303G      & 66.6\%      & 60.3\%      \\
\dag~VideoSwin\cite{Liu2022-videoswin} & CVPR'22 & VideoSwin-T & 27.9M &478G      & 64.3\%      & 60.6\%      \\
\arrayrulecolor{gray}\midrule
\multicolumn{3}{l}{\textbf{Supervised GAR}} \\
\addlinespace[0.5ex]
\dag~ARG\cite{Wu2019-ARG} & CVPR'19 & Resnet-18 & 49.5M &307G      & 59.0\%      & 56.8\%      \\
\dag~AT\cite{Gavrilyuk2020-ATGAR} & CVPR'20 & Resnet-18 & 29.6M &305G      & 47.1\%      & 41.5\%      \\
\dag~SACRF\cite{Pramono2020-SACRF} & ECCV'20 & Resnet-18 & 53.7M &339G      & 56.3\%      & 52.8\%      \\
\dag~DIN\cite{Yuan2021-DIN} & AAAI'21 & Resnet-18 & 26.0M &304G      & 61.6\%      & 56.0\%      \\
KRG-GAR~\cite{Pei2023-keyrole} & TCSVT'23 & Resnet-18 & - & - & 72.4\% & 67.1\% \\
\arrayrulecolor{gray}\midrule
\multicolumn{3}{l}{\textbf{Weakly supervised GAR}} \\
\addlinespace[0.5ex]
SAM~\cite{Yan2020-SAM} & ECCV'20 & Resnet-18 & -             &-         & 49.1\%      & 47.5\%      \\
\dag~SAM~\cite{Yan2020-SAM} & ECCV'20 & Resnet-18 & 25.5M &304G      & 54.3\%      & 51.5\%      \\
Dual-AI (RGB)~\cite{Han2022-dualAI} & CVPR'22 & Inception-v3 & - & - & 58.1\% & 50.2\% \\
DFWSGAR~\cite{Kim2022-dfwsgar} & CVPR'22 & Resnet-18 & 17.5M &313G      & 75.8\%     & 71.2\%     \\
${\ast}$~GSTCo-PCE~\cite{Du2023-GSTCO} & TCSVT'23 & Resnet-18 & - & -      & 68.6\%     & 64.2\%     \\
LRMM+GCM~\cite{Du2023-LRMMGCM} & ImaVis'23 & Resnet-18 & 14.2M &306G      & 77.8\%     & \underline{73.2\%}     \\
${\ast}$~SPARTAN\cite{Chappa2023-SPARTAN} & CVPRW'23 & ViT-Base & - & - & \textbf{82.1\%} & 72.8\% \\
\rowcolor{Gray}
\netnameshort(Ours) & -  & Resnet-18 & 13.8M &307G    & \underline{79.1\%}     & \textbf{76.0\%}     \\
\hline
\end{tabular}
 }
\end{center}
\vspace{-3em}
\end{table}

\subsection{Comparison with State-of-the-art}\vspace{\subsecmargin}
We compare our method with the state-of-the-art methods in GAR\cite{Wu2019-ARG,Gavrilyuk2020-ATGAR,Pramono2020-SACRF,Pei2023-keyrole} and WSGAR~\cite{Yan2020-SAM,Kim2022-dfwsgar,Du2023-LRMMGCM,Chappa2023-SPARTAN} on the NBA and Volleyball dataset. \Cref{table:SOTA_NBA} summarizes the NBA dataset results. NBA dataset does not provide ground truth bounding boxes, so we use the bounding box proposals provided by SAM~\cite{Yan2020-SAM} to accommodate fully supervised learning if needed. 
%
Only the RGB frames were available during the inference stage. Among methods that require only one end-to-end training stage, \netnameshort~performs best with 1.3\%p higher MCA and 2.8\%p higher MPCA. SPARTAN\cite{Chappa2023-SPARTAN} achieved higher MCA, but it requires additional training stages due to pretraining and a different backbone of the video transformer network. We also achieved a higher MPCA of 3.2\%p compared to SPARTAN, which shows the better ability of \netnameshort~to tackle imbalanced classes. Our model also requires a smaller parameter size with a similar number of FLOPs compared with the lowest one among the reported models. Our performance also surpassed the recent video backbones, ResNet-18 TSM~\cite{Lin2019-tsm} and VideoSwin-T~\cite{Liu2022-videoswin}, used in conventional action recognition studies. Our model also performed competitively in the Volleyball dataset as shown in \cref{table:SOTA_Volley}, surpassing several fully-supervised and all weakly-supervised methods in terms of the 8-classes MCA. The Volleyball dataset contains fewer long-range activity sequences compared to the NBA dataset. The higher difference of \netnameshort~with other methods on the NBA dataset shows that it is better suited for long-range activity sequences with more complex inter-frame inter-actor relationships.

\begin{table}[t]
\begin{center}
\caption{\textbf{Comparison with the state-of-the-art GAR models and video backbones on the Volleyball~\cite{Ibrahim2016-volleyball} dataset}. Numbers in \textbf{bold} indicate the best performance among WSGAR. '\dag' indicates that the results are from Kim~\etal\cite{Kim2022-dfwsgar} reproduction, and others are as reported from the original paper. ${\ast}$ indicates the method requires additional pre-training before a supervised fine-tuning.
}
\label{table:SOTA_Volley}
\vspace{-1em}
\resizebox{0.8\columnwidth}{!}{
\setlength{\tabcolsep}{2mm}
\begin{tabular}{l c c c c}
\hline
Method & Published on & Backbone & MCA{\up} & Merged MCA{\up}      \\
\hline
\addlinespace[0.5ex]
\multicolumn{3}{l}{\textbf{Supervised GAR}} \\
\addlinespace[0.5ex]
\dag~PCTDM\cite{yan2018participation} & ACMMM'18 & ResNet-18     &90.3\%      & 94.3\%      \\

\dag~ARG\cite{Wu2019-ARG} & CVPR'19 & ResNet-18    & 91.1\%      &  95.1\%      \\
\dag~AT\cite{Gavrilyuk2020-ATGAR} & CVPR'20 &  ResNet-18     &  90.0\%      & 94.0\%      \\
\dag~SACRF\cite{Pramono2020-SACRF} & ECCV'20 &  ResNet-18     & 90.7\%      &92.7\%      \\
\dag~DIN\cite{Yuan2021-DIN} & ICCV'21 & ResNet-18    &  93.1\%      & 95.6\%      \\

GroupFormer~\cite{Li2021-Groupformer} & ICCV'21 & Inception-v3 & 94.1\% & - \\
Dual-AI (RGB)~\cite{Han2022-dualAI} & CVPR'22 & Inception-v3 & 94.4\% & - \\
KRG-GAR~\cite{Pei2023-keyrole} & TCSVT'23 & Inception-v3 & 95.4\% & -   \\
\arrayrulecolor{gray}\midrule
\multicolumn{3}{l}{\textbf{Weakly supervised GAR}} \\
\addlinespace[0.5ex]
SAM~\cite{Yan2020-SAM} & ECCV'20 & ResNet-18        & 86.3\%      & 93.1\%      \\
\dag~SAM~\cite{Yan2020-SAM} & ECCV'20 & Inception-v3     &-      &  94.0\%      \\
DFWSGAR~\cite{Kim2022-dfwsgar} & CVPR'22 & ResNet-18       & 90.5\%     & 94.4\%     \\

Dual-AI (RGB)~\cite{Han2022-dualAI} & CVPR'22 & Inception-v3 & - & \textbf{95.8\%} \\
LRMM+GCM~\cite{Du2023-LRMMGCM} & ImaVis'23 &ResNet-18      & 92.8\%     & 95.6\%   \\
${\ast}$~SPARTAN~\cite{Chappa2023-SPARTAN} & CVPRW'23 &ViT-Base & 92.9\% & 95.6\% \\
Wu~\etal~\cite{Wu2023-active} & TCSVT'23 & Inception-v3 & 90.2\% & 94.9\% \\
\rowcolor{Gray} \netnameshort(Ours) & - & ResNet-18    & 91.6\% & 94.5\%     \\
\rowcolor{Gray} \netnameshort(Ours) & - & Inception-v3    & \textbf{93.3\%} & 95.2\%     \\
\hline
\end{tabular}
}
\end{center}
\vspace{-5mm}
\end{table}

\begin{table}[t]
\captionof{table}{\textbf{Ablation on actor relation module.}}
\label{tab:abl_loss4}
\vspace{-3mm}
\centering
  \setlength{\tabcolsep}{5mm}{
  \renewcommand\arraystretch{0.8}
  \resizebox{0.75\columnwidth}{!}{
  \begin{tabular}{l|cc}
  \toprule
  Actor relation module & MCA & MPCA \\
  \midrule
  Actor motion path &\ 74.3\% & 73.9\% \\
  Actor motion path + Aux. losses &\ 77.1\% & 71.7\% \\
  Mirror dual path  &\ 77.0\% & 70.7\% \\
  \rowcolor{Gray}\netnameshort&\ 79.1\% & 76.0\% \\
    \bottomrule
  \end{tabular}}}
  \vspace{-5mm}
\end{table}

\subsection{Ablations}\vspace{\subsecmargin}
\subsubsection{Spatio-temporal relationship between actors.} \Cref{tab:abl_loss4} shows the result of experimenting methods to construct the spatio-temporal relationships among the actor features. First, we set the baseline of using only the actor motion path. In the second row, we show the effectiveness of our proposed additional losses ($\mathcal{L}_{\text{gf}},\mathcal{L}_{\text{tco}}, \mathcal{L}_{\text{flm}}$) as we obtained an increase in MCA. In the third experiment, we added a dual path that is set as a mirror of the actor-centric path, i.e. simply reversed the order of spatial and temporal processing, resulting in no improvement. Lastly, we experimented with the \netnameshort~strategy that adds the group-centric path combining actor-relation MHSA and 2D spatio-temporal convolution. The high performance proves the effectiveness of our strategy in building complementing actor-motion and group-motion representation.

\begin{figure}[t]
\begin{minipage}{\linewidth}
\begin{minipage}[t]{0.27\linewidth}
\captionof{table}{\textbf{Ablation on the number of $K_{\text{flw}}$.}}
\label{tab:ablation_num_token}
\begin{tabular}[c]{>{\centering\arraybackslash}m{0.8cm}>{\centering\arraybackslash}m{1.1cm}>{\centering\arraybackslash}m{1.1cm}}
    \toprule
    $K_{\text{flm}}$    & MCA               & MPCA              \\
    \midrule
    4    & 78.2\%    & 72.6\%    \\
    \rowcolor{Gray}6  & 79.1\%  & 76.0\%              \\
    9       & 78.2\%  & 72.9\%  \\
    12  & 77.2\%    & 69.9\%   \\
    \bottomrule
\end{tabular}
\end{minipage}
\hfill
\begin{minipage}[t]{0.33\linewidth}
\captionof{table}{\textbf{Ablation on the number of aligned attention map encoder blocks.}}
\label{tab:ablation_layer_flow}
\begin{tabular}[c]{>{\centering\arraybackslash}m{1.3cm}>{\centering\arraybackslash}m{1.2cm}>{\centering\arraybackslash}m{1.2cm}}
    \toprule
    $L_{\text{flm}}$  & MCA & MPCA  \\
    \midrule
    $1^{\text{st}}$ Half  & 77.2\%                  & 71.1\%  \\
    $2^{\text{nd}}$ Half  & 77.9\%  & 71.9\%  \\
    \rowcolor{Gray}All & 79.1\% & 76.0\%  \\
    \bottomrule
\end{tabular}
\end{minipage}
\hfill
\begin{minipage}[t]{0.35\linewidth}
\captionof{table}{\textbf{Ablation on alignment auxiliary losses.}}
\label{tab:ablation_align_loss}
\begin{tabular}[c]{>{\centering\arraybackslash}m{1.5cm}>{\centering\arraybackslash}m{1.2cm}>{\centering\arraybackslash}m{1.2cm}}
  \toprule
  Loss Formula & MCA & MPCA \\
  \midrule
  L1 $\mathcal{L}_{\text{tco}}$ &\ 76.8\% & 72.2\% \\
  L1 $\mathcal{L}_{\text{flm}}$ &\ 76.2\% & 73.3\% \\
  \rowcolor{Gray}Contra. &\ 79.1\% & 76.0\% \\
  \bottomrule
\end{tabular}
\end{minipage}
\end{minipage}
\vspace{-4mm}
\end{figure}

\vspace{\subsubsecmargin}\subsubsection{Effect of $K_{\text{flw}}$.} We varied the number of actor attention token maps averaged and aligned to the flow map in \cref{tab:ablation_num_token}. The best result of a balanced number of $K_{\text{flw}}$ shows the need to maintain a balance between using optical flow to find key actors and using other factors.

\vspace{\subsubsecmargin}\subsubsection{Multi-level attention map alignment}. In \cref{eq:loss_flow}, we align every level of the motion-aware encoder block to the flow map. For the ablation study in \cref{tab:ablation_layer_flow}, we varied the level of blocks aligned with the flow map. There is a slight improvement when aligning only the second half of blocks compared to only the first half of blocks. We achieved further improvement once we aligned all the blocks, showing all encoder blocks benefit from the alignment process.

\vspace{\subsubsecmargin}\subsubsection{Alignment function.}
We use the contrastive loss in \cref{eq:loss_flow} to align the attention map with the optical flow map and in \cref{eq:loss_alignment_2} to align the temporally adjacent actor tokens with the same index. We alternatively experimented using L1 loss function to calculate the $\ell_{\mbox{\scriptsize flm}}$ and $\ell_{\mbox{\scriptsize tco}}$ value, with results shown in \cref{tab:ablation_align_loss}. Changing either of the contrastive losses to L1 loss results in a lower performance score, thus justifying the usage of contrastive loss in our formulation.

\begin{figure}[t]
\vspace{-2.5mm}
\begin{minipage}{\linewidth}
\begin{minipage}[t]{0.58\linewidth}
\captionof{table}{\textbf{Ablation study on training losses.}}
\label{tab:abl_loss}
\setlength{\tabcolsep}{2mm}
  \renewcommand\arraystretch{0.9}
  \resizebox{\columnwidth}{!}{
  \begin{tabular}{c|ccc|cc}
  \toprule
  No & $\mathcal{L}_{\text{gf}}$ & $\mathcal{L}_{\text{tco}}$ & $\mathcal{L}_{\text{flm}}$ & MCA & MPCA \\
  \midrule
  1 & \textcolor{black}{\xmark} & \textcolor{black}{\xmark} & \textcolor{black}{\xmark}  &\ 75.3\% & 70.1\% \\
  2 & \textcolor{black}{\cmark} & \textcolor{black}{\xmark} & \textcolor{black}{\xmark}  &\ 78.3\% & 72.3\% \\
  3 & \textcolor{black}{\cmark} & \textcolor{black}{\cmark} & \textcolor{black}{\xmark}  &\ 77.5\% & 73.0\% \\
4 & \textcolor{black}{\cmark} & \textcolor{black}{\xmark} & \textcolor{black}{\cmark}  &\ 77.5\% & 73.0\% \\
  \rowcolor{Gray}5 & \textcolor{black}{\cmark}& \textcolor{black}{\cmark} & \textcolor{black}{\cmark} &  \ 79.1\% & 76.0\% \\
  \bottomrule
  \end{tabular}}
\end{minipage}
\hfill
\begin{minipage}[t]{0.39\linewidth}
\captionof{table}{\textbf{Ablation on detaching loss from spatio-temporal convolution module.}}
\label{tab:ablation_detach_loss}
\setlength{\tabcolsep}{0.8mm}
\renewcommand\arraystretch{0.9}
 \resizebox{0.8\columnwidth}{!}{
 \begin{tabular}{c|c c}
  \toprule
  Detach & \multirow{2}{*}{MCA} & \multirow{2}{*}{MPCA}\\ loss & & \\
  \midrule
  \xmark  & 77.3\%  & 72.5\%  \\
  \rowcolor{Gray} \cmark  & 79.1\% & 76.0\% \\
  \bottomrule
  \end{tabular}
  }
\end{minipage}

\end{minipage}
\vspace{-8mm}
\end{figure}

\vspace{\subsubsecmargin}\subsubsection{Effect of auxiliary losses.} We validated the effect our auxiliary losses in \cref{tab:abl_loss}. Firstly, we used only the activity cross-entropy loss as the learning function, which results in suboptimal performance. Then, we added the frame-level classifier loss $\mathcal{L}_{\text{gf}}$ which increases the performance as the group motion path properly functions. Experiments 3 and 4 show that if the motion-aware actor encoder auxiliary losses $L_{\text {tco}}$ and $L_{\text{flm}}$ are added individually, they can be contra-productive. But, when combined as in Exp. 5, they improved the model performance.

\vspace{\subsubsecmargin}\subsubsection{Detached loss before spatiotemporal convolution.} 
We compared the model learning effectiveness between attaching the loss from the spatio-temporal convolution module in the group motion path and detaching them. Results in \cref{tab:ablation_detach_loss} show better performance by detaching the loss.

\begin{figure*}[t]
\begin{center}
\includegraphics[width=\linewidth]{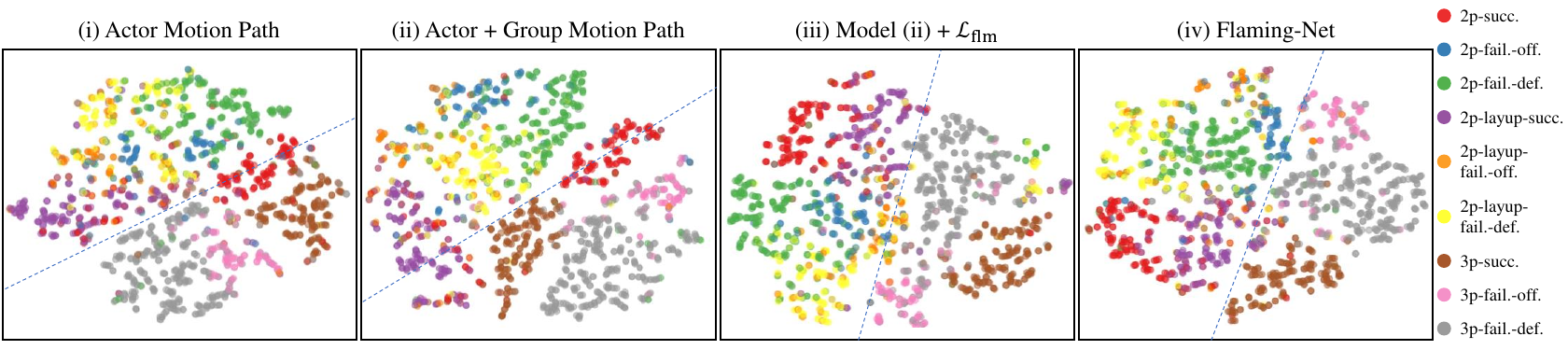}
\end{center}
\vspace{-7mm}
\caption{\textbf{t-SNE~\cite{van2008-tsne} visualization} of feature embedding learned by different model variants on the NBA~\cite{Yan2020-SAM} dataset.}
\label{fig:fig6}
\vspace{-1.0em}
\end{figure*}

\begin{figure}[t]
\vspace{-1mm}
  \centering
  \includegraphics[width=\linewidth]{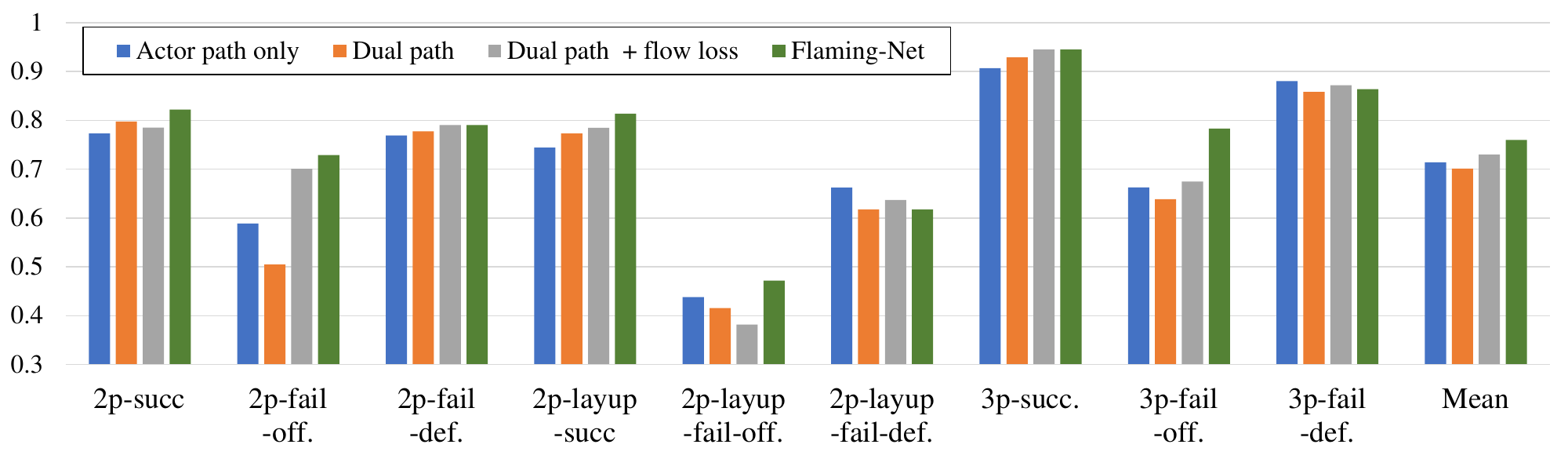}
   \vspace{-8mm}
   \caption{\textbf{Per-class accuracy on NBA~\cite{Yan2020-SAM}} dataset across model variants.}
   \label{fig:fig_classes}
\vspace{-8mm}
\end{figure}

\subsection{Qualitative Analysis}\vspace{\subsecmargin}
\Cref{fig:fig6} displays the t-SNE\cite{van2008-tsne} visualization result, where the final group representation of each model on NBA is visualized in two-dimensional space, of 4 model variants; (i) model with only the actor motion path, (ii) model with actor motion path and group motion path along with $\mathcal{L_{\text{gf}}}$, (iii) model (ii) with additional ${\mathcal{L}_{\text{flm}}}$, and (iv) complete \netnameshort~model. To help understand the impact of Flaming-Net in projecting features between activities, we draw a boundary line that divides the area into 3-point related activities to the right side of the line, and other activities on the opposite side. As \netnameshort~components are added, the activities are better separated, i.e. less amount of misclassified non-3-point activities to the right of the line. Interestingly, it also puts activities related to successful attempts nearly together. This shows the model understands the visual context after scoring a point is similar for these activities. 

\begin{figure}[t]
  \centering
  \includegraphics[width=0.95\linewidth]{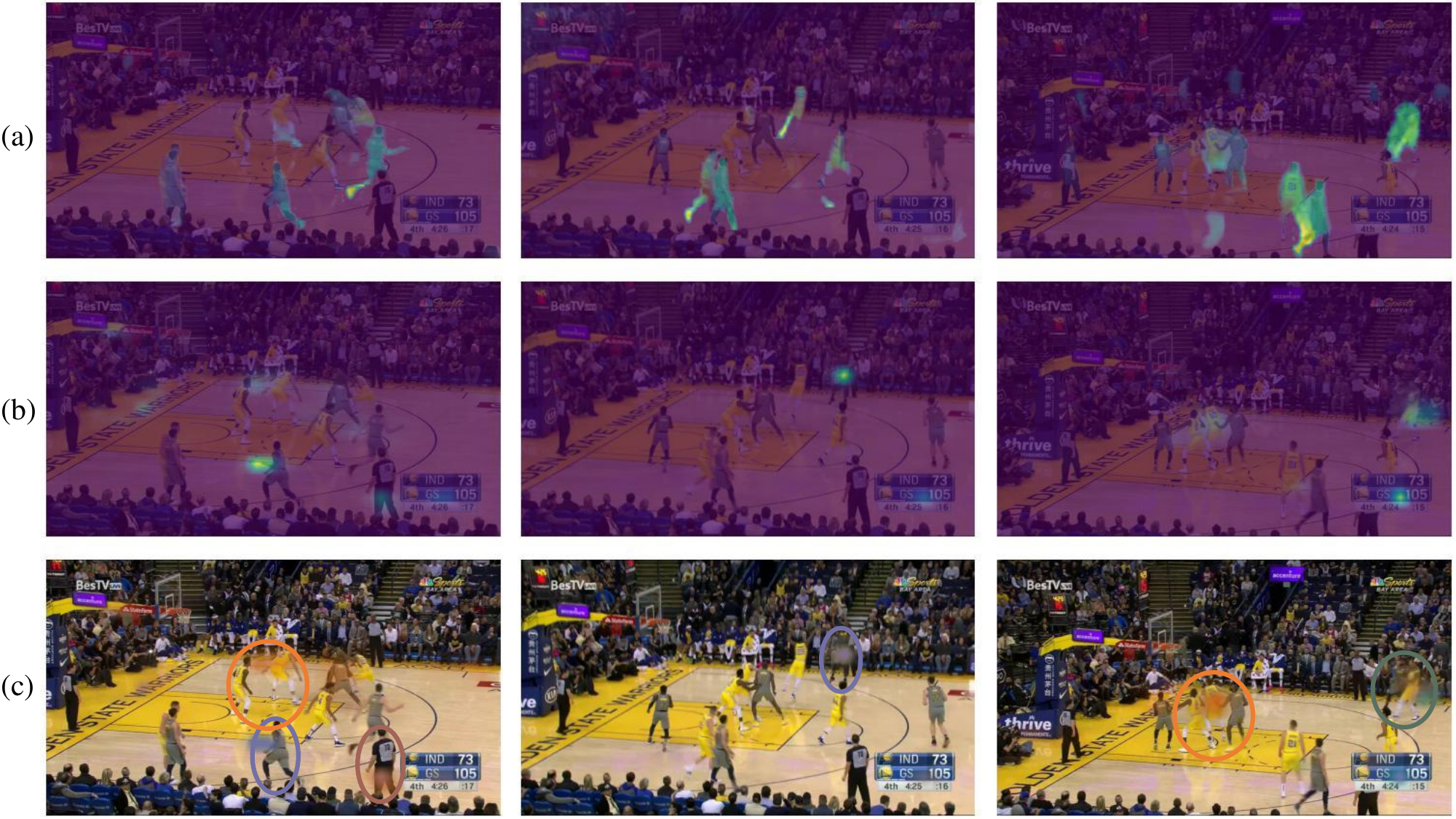}
   \vspace{-3mm}
   \caption{\textbf{Token attention visualization.} In the first row, we show RGB frames overlayed with the flow map. In the second row, we overlay the RGB frames with the encoder attention map. In the third row, we display separately the attention value of three different tokens, and also draw a circle on actors related to each token, to show how our encoder uses different tokens for different actors.}
   \label{fig:fig_tokens}
\vspace{-6mm}
\end{figure}

\Cref{fig:fig_classes} shows the per-class accuracy of \netnameshort~variants. Without auxiliary losses and flow guidance, the dual path (actor motion path + group motion path) underperforms the actor-centric path-only variant. Adding the flow-guided loss and other auxiliary losses enhances the performance, particularly for the activities related to offensive rebounds and successful scorings, achieving state-of-the-art performance. \Cref{fig:fig_tokens} displays the comparison between the flow map and the attention of the actor encoder. Comparing the flow map (a) with the attention map (b), the model selectively follows the flow guide if the actor is useful to the activity and tends to ignore the influence of irrelevant flow. The third row (c) visualizes separate token attention maps differentiated by colors, and we also circle the actors represented by the tokens. Our model has diverse attention with each token representing different relevant actors to the activity.
\section{Conclusion}
\vspace{\subsecmargin}
We have introduced the \netnameshort~to address the challenge of WSGAR. The optical flow map is used as learning guidance to enhance the motion awareness of the actor encoder when extracting key actors. Then, we construct a comprehensive spatio-temporal dual-path inference module that combines the strengths of building group representation from relations among multiple individual actor motions and direct inter-actor inter-frame connections. \netnameshort~achieves state-of-the-art performance on two GAR datasets. Ablation studies validate the effectiveness of our learning strategy and the actor-relation module aggregation. 

\vspace{\subsubsecmargin}\subsubsection{Discussion.} In future work, we aim to alleviate the limitations of our current approach, such as removing the need for an external optical map generation network and minimizing the number of network hyperparameters. Besides that, there are multiple interesting topics to expand our GAR solutions. For example, increasing efficiency for the feasibility of real-time or online scenarios or adding an explainable inference to inform the user explicitly on how the actors interact with each other and their role in the activity.

\clearpage  

%
%
\bibliographystyle{splncs04}
\bibliography{main}

\end{document}